\documentclass[letterpaper]{article} 
\usepackage{aaai25}  
\usepackage{times}  
\usepackage{helvet}  
\usepackage{courier}  
\usepackage[hyphens]{url}  
\usepackage{graphicx} 
\usepackage{natbib}  
\usepackage{caption} 
\usepackage{algorithm}
\usepackage{algorithmic}

\usepackage{amsmath}
\usepackage{multirow}
\usepackage{makecell}
\usepackage{pifont}
\usepackage{newfloat}
\usepackage{listings}
\DeclareCaptionStyle{ruled}{labelfont=normalfont,labelsep=colon,strut=off} 
\floatstyle{ruled}
\newfloat{listing}{tb}{lst}{}
\floatname{listing}{Listing}
\title{Efficient Indoor Depth Completion Network using \\ Mask-adaptive Gated Convolution}

\author{
    Tingxuan Huang\textsuperscript{\rm 1},
    Jiacheng Miao\textsuperscript{\rm 1},
    Shizhuo Deng\textsuperscript{\rm 1, \rm 2},
    Tong Jia\textsuperscript{\rm 1, \rm 2},
    Dongyue Chen\textsuperscript{\rm 1, \rm 2, \rm3, \footnote{Corresponding Author}}\\
}
\affiliations{
    \textsuperscript{\rm 1}College of Information Science and Engineering, Northeastern University, China\\
    \textsuperscript{\rm 2}Foshan Graduate School of Innovation, Northeastern University, China\\
    \textsuperscript{\rm 3}National Frontiers Science Center for Industrial Intelligence and Systems Optimization, Northeastern University, China\\

    \{cangshuhuang, mjc24500817\}@gmail.com, dengshizhuo@mail.neu.edu.cn, \{jiatong, chendongyue\}@ise.neu.edu.cn
}
\usepackage{bibentry}

\begin{document}

\maketitle

\begin{abstract}
Most indoor depth completion tasks rely on convolutional auto-encoders to reconstruct depth images, especially in areas with significant missing values. While traditional convolution treats valid and missing pixels equally, Partial Convolution (PConv) has mitigated this limitation. However, PConv fails to distinguish the varying degree of invalidity across different missing areas, which highlights the need for a more refined strategy. To solve this problem, we propose a novel system for indoor depth completion tasks that leverages Mask-adaptive Gated Convolution (MagaConv). MagaConv utilizes gated signals to selectively apply convolution kernels based on the characteristics of missing depth data. These gating signals are generated using shared convolution kernels that jointly process depth features and corresponding masks, ensuring coherent weight optimization. Additionally, the mask undergoes iterative updates according to predefined rules. To improve the fusion of depth and color information, we introduce a Bi-directional Aligning Projection (Bid-AP) module, which utilizes a bi-directional projection scheme with global spatial-channel attention mechanisms to filter out depth-irrelevant features from other modalities. Extensive experiments on popular benchmarks, including NYU-Depth V2, DIML, and SUN RGB-D, demonstrate that our model outperforms state-of-the-art methods in both accuracy and efficiency. The code is available at \textit{https://github.com/htx0601/MagaConv}.
\end{abstract}

\section{Introduction}

Depth completion, or depth inpainting, is vital for filling missing pixels in depth images, crucial for applications such as 3D reconstruction \cite{3d}, virtual reality \cite{vr}, and autonomous vehicles \cite{auto-driving}. It aims to efficiently replace missing pixels in raw depth maps acquired from sensors like Time-of-flight, structured light, Lidar, and binocular vision. In indoor environments, inherent limitations of these sensors, such as sensor noise, reflections, absorption, or sharp boundaries often result in incomplete data. Overcoming these challenges and developing robust depth completion algorithms is essential for obtaining accurate depth maps. 

Recent methods use encoder-decoder architectures like U-Net and its variants \cite{unet} to predict depth. However, the vanilla convolution, which treats all pixels equally, including missing ones, can lead to inaccuracies and error propagation in neighboring regions. Approaches like dilated convolutions \cite{dilated}, partial convolutions \cite{partial}, gated convolutions \cite{gated}, and attention-guided gated convolutions \cite{aggnet} aim to improve accuracy by handling missing data by adjusting kernel positions or suppressing invalid features related to missing pixels. However, they have not fully exploited the potential impact of the invalid pixel in extracting depth features.

\begin{figure}[t]
\centering
\includegraphics[width=\linewidth]{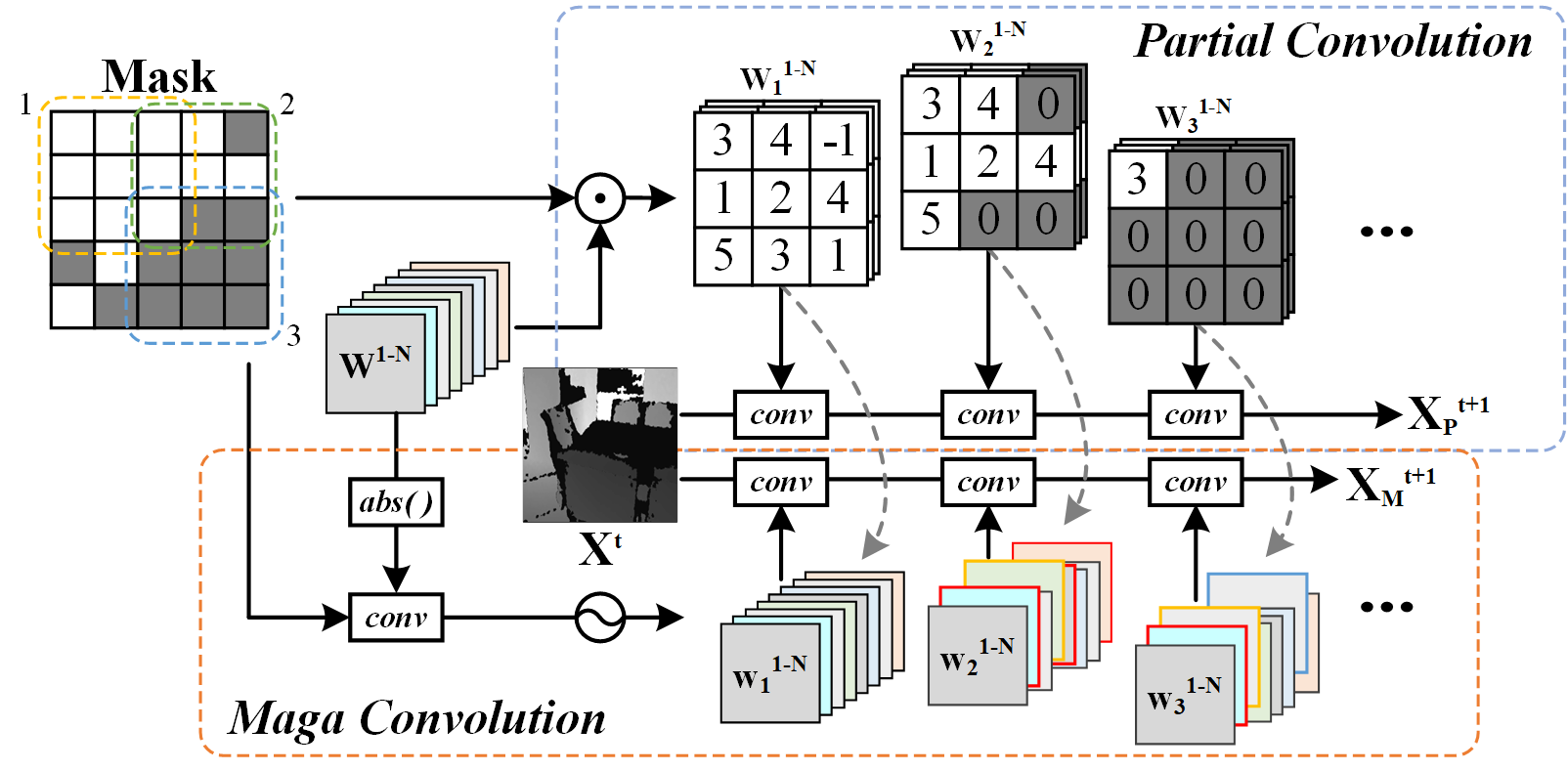}
\caption{The comparison between Partial Convolution and Maga Convolution, designed to encode incomplete depth images using associated masks. Here, $X^t$ is the input/output feature at encoding step $t$. $W_i$ is the specific convolution kernels applied at position $i$. While PConv ensures output from valid pixels, it overlooks the challenge of using the same kernels for various invalidity levels, as it may mask out crucial parameters in $W$. MagaConv addresses this by selecting kernels tailored to specific invalid patterns.}
\label{fig:fig1}
\end{figure}

Take the Partial Convolution (PConv) as an example, as shown in Fig.  \ref{fig:fig1}. It employs a binary mask to distinguish between valid and invalid data during convolution. Throughout each convolutional operation, this mask interacts with input features across all layers, ensuring that the resulting outputs only derive from valid pixels, thus guaranteeing reliability. However, this reliability assumption has two key limitations. Firstly, the convolutional receptive field contains varying numbers of invalid pixels with diverse distributions, and simply discarding these pixels overlooks crucial details. Secondly, employing identical convolution kernels across different invalid contexts lacks adaptability. The convolution kernel's parameters are intended to capture crucial features and patterns. Partially masking key parameters disrupts the learned features, preventing the full utilization of learned information and leading to unreliability.

We introduce Mask-adaptive Gated Convolution (MagaConv), a novel convolutional operation modulated by iteratively updated masks to solve these challenges. It enhances depth feature extraction by selecting convolution kernels based on the specific characteristics of incomplete depth data. It is achieved by dynamically generating gating signals to evaluate each convolutional operation. By employing shared convolution kernels that process both depth features and corresponding masks, MagaConv can determine the degree of invalidity at each position within every channel. This information is then converted into a gating signal, through a unique activation function, to selectively choose kernels in a manner that prevents disruption of their essential parameters. Additionally, MagaConv iteratively updates masks to gradually complete depth features, effectively filling large holes and enabling precise extraction.

After depth coarsely completion and encoding, the next step involves integrating them with color information and decoding. Researchers have explored various RGB-guided approaches \cite{cspn, cheng2020cspn++, completionformer} that typically fuse features by concatenating them and applying standard convolutions. However, these approaches face limitations. Firstly, they neglect the differences between color and depth modalities: depth captures geometry, while RGB depicts appearance and texture \cite{aggnet}. Simply concatenating risks introduces depth-irrelevant features and misses complementary information. Secondly, localized convolution operations fail to capture the global context, crucial for understanding spatial relationships between distant objects.

To tackle these issues, we have considered using transformers for cross-attention mechanisms as a potential remedy \cite{transformer}. However, due to the limited availability of labeled indoor RGB-D pairs in most public datasets, transformers may struggle to learn the complex relationship between the two modalities \cite{liu2021efficient}. Therefore, inspired by spatial-adaptive normalization \cite{spade}, we introduce a novel module named Bi-directional Aligning Projection (Bid-AP), facilitating a comprehensive alignment of these modalities from a global perspective.

In general, our main contribution can be summarized as follows:
\begin{itemize}
\item We develop an efficient convolutional encoder-decoder network that utilizes our newly proposed MagaConv and Bid-AP to generate high-quality completion of the indoor depth image.
\item A Mask-adaptive Gated Convolution (MagaConv) is proposed to extract reliable depth features while considering the degree of invalidity in missing regions. MagaConv utilizes a shared convolution operation and iteratively updated masks to modulate the encoding process.
\item A Bi-directional Aligning Projection module (Bid-AP) is proposed, leveraging MLP-based spatial-adaptive normalization to align with color data while filtering out depth-irrelevant features.
\item Experimental results demonstrate that our model outperforms the state-of-the-art on three popular benchmarks, including NYU-Depth V2, DIML, and SUN RGB-D datasets.
\end{itemize}

\section{Related Works}
\label{sec:related works}
\subsection{Depth Completion}

The task of depth completion aims to generate dense depth maps from incomplete depth images. \cite{sparse2depth, qu2020depth, wang2023lrru, Yanaaai, yanicml, wang2024improving} employed encoder-decoder networks to obtain dense depth maps. S2DNet and Deepdnet \cite{hambarde2020s2dnet, hegde2021deepdnet} proposed a two-stage network, focusing on acquiring approximate depth images and enhancing the primary results.\cite{gu2021denselidar, related4, zhu2022robust} introduced residual depth map completion networks, which utilize residual maps to enhance the initial completion image, resulting in sharper edges. SPN, CSPN, CSPN++, NLSPN, GraphSPN, DySPN \cite{liu2017learning, cheng2018depth, cheng2020cspn++, NLSPN, graphcspn, dyspn} optimized the SPN algorithms to enhance the prediction of unfamiliar depth values by effectively incorporating known depth information. However, the predicted depth maps still exhibit blurriness attributed to the limitations of vanilla convolution operations in encoding depth features.

\begin{figure*}[t]
\centering
\includegraphics[width=0.75\linewidth]{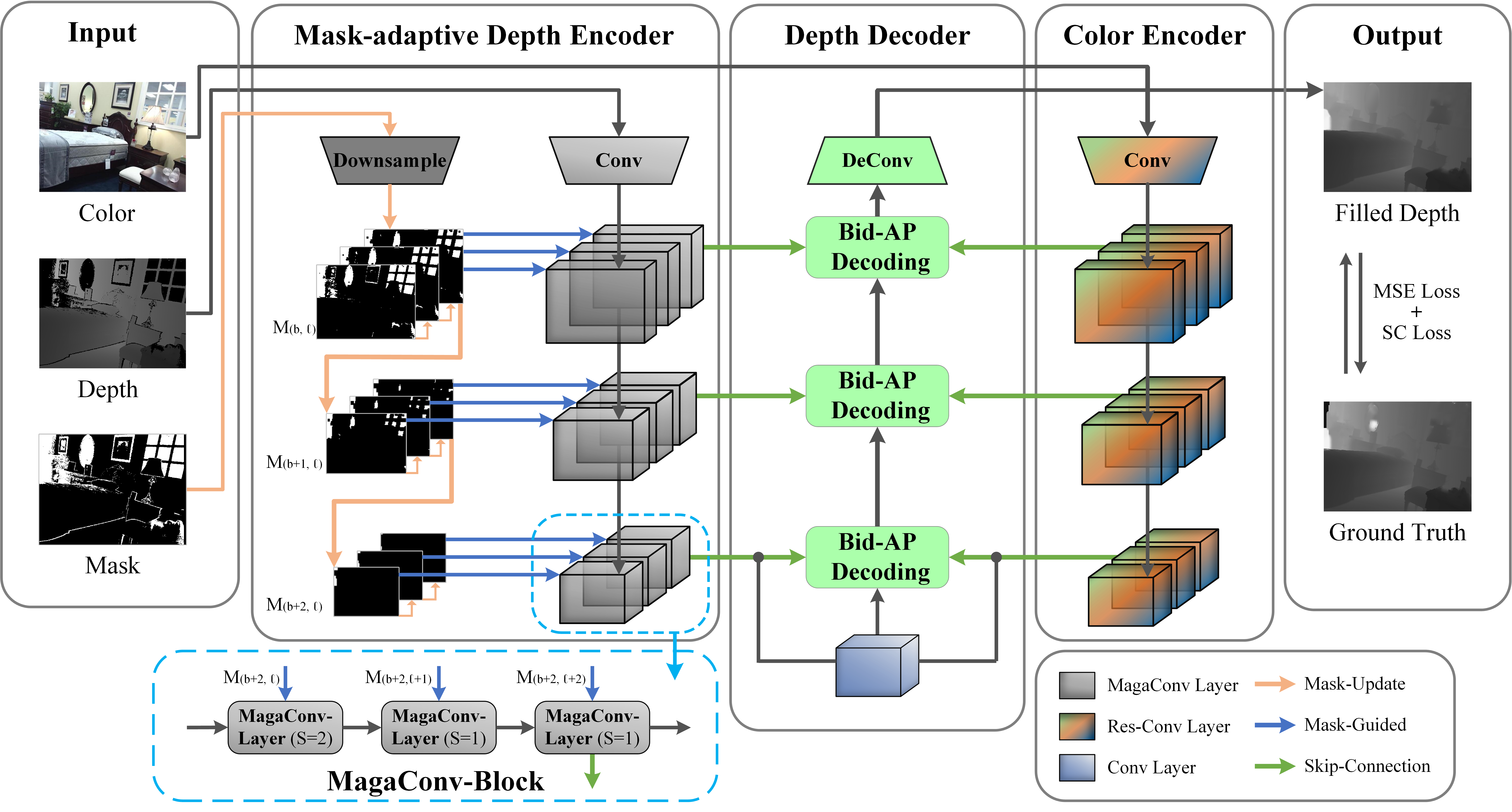}
\caption{Pipeline of our depth completion model, including the MagaConv architecture, the Bid-AP module we proposed. $M_{(b, l)}$ represents adaptive masks, where $b$ and $l$ represent the block and layer, respectively.}
\label{fig:net}
\end{figure*}

\subsection{Feature Extraction}

The presence of missing or unreliable depth pixels in the raw depth map poses a challenge when using VConv to extract features. To overcome the artifacts, researchers have proposed a series of convolutions \cite{partial, gated, FFC, LBAM} to avoid the impact of missing value. Besides PConv, \cite{gated} introduced a Gated Convolution (GConv) that generalizes partial convolution by providing a learnable dynamic feature selection mechanism. \cite{FFC} proposed Fast Fourier Convolution, which has a larger receptive field and a cross-scale fusion within the convolution. \cite{LBAM} introduced a Learnable Bidirectional Attention Maps (LBAM) module that learns feature re-normalization and mask-updating in an end-to-end manner. Nevertheless, these methods did not adequately exploit the use of masks that mark that marks the invalid pixels. To tackle this issue, we propose the MagaConv, a new convolution operation that is modulated by iteratively updated masks.

\subsection{Multi-modal Data Fusion}

Multi-modal feature fusion is another essential aspect compared to depth feature extraction. \cite{xu2019depth, imran2019depth} used direct channel-wise concatenation to fuse features. \cite{zhong2019deep, li2020multi, zhou2022self, yaneccv} designed adaptive modules to realize global feature fusion throughout the encoding and decoding procedures.\cite{aggnet} proposed Attention Guided Gated-Convolution (AG-GConv) to fuse depth and color features at different scales, effectively reducing the negative impacts of invalid depth data on the reconstruction. Additionally, Rignet, GuideNet, and Ssgp \cite{rignet, guidenet, schuster2021ssgp, tang2024bilateral} adopt dual-modal encoder-decoder networks, enabling a multi-level fusion within the network architecture. In this paper, We propose the Bi-directional Aligning Projection (Bid-AP) module, which aims to comprehensively align depth-relevant cues from the two modalities, and fuse the features in a global perspective.

\section{Methods}

In this section, we present our overall depth completion network architecture and its two key components: Mask-adaptive Gated Convolution (MagaConv) and Bi-directional Aligning Projection (Bid-AP). Additionally, we introduce the overall loss function, including an MSE loss and a Structure-Consistency loss.

\subsection{Overall Network Architecture}

\label{subsec:overall}
The pipeline of our model is shown in Fig. \ref{fig:net}, it aims to fill all the missing depth pixels in raw depth images with the guidance of color images. The network consists of three components: (i) Mask-adaptive Depth Encoder, (ii) Color Encoder, and (iii) Depth Decoder with Bid-AP decoding layer. 

(i) The Mask-adaptive Depth Encoder is designed to extract reliable depth features while addressing missing data issues. The encoding procedure operates in three levels: MagaConv-Blocks (M-Blocks), MagaConv-Layers (M-Layers), and MagaConv. The encoder consists of three M-Blocks, each of which downsamples the depth feature by half. Within each block, input data undergoes three sequential M-Layers, each steered by distinct masks that are refreshed per block and layer. The initial layer strides by 2, while the subsequent layers use a stride of 1. Only the output features from the final layer of each block are then forwarded to the decoder via skip connections.

(ii) The Color Encoder takes RGB images as inputs to extract depth-relevant features. Its architecture mirrors that of the Depth Encoder, with three convolutional layers in each block. However, it utilizes standard convolutional layers instead of MagaConv, integrates residual connections, and does not rely on pre-trained parameters.

(iii) The Depth Decoder, enhanced with the Bid-AP module, leverages multi-scale, skip-connected pathways to reconstruct a complete depth image. The Bid-AP module integrates features from both encoders, ensuring thorough alignment and capturing complementary details. By applying the Bid-AP module across different scales, the Depth Decoder is capable of generating high-quality depth completion results.

\subsection{Mask-adaptive Gated Convolution}
\label{subsec:magaconv}

\begin{figure}[t]
    \centering
  \includegraphics[width=1\linewidth]{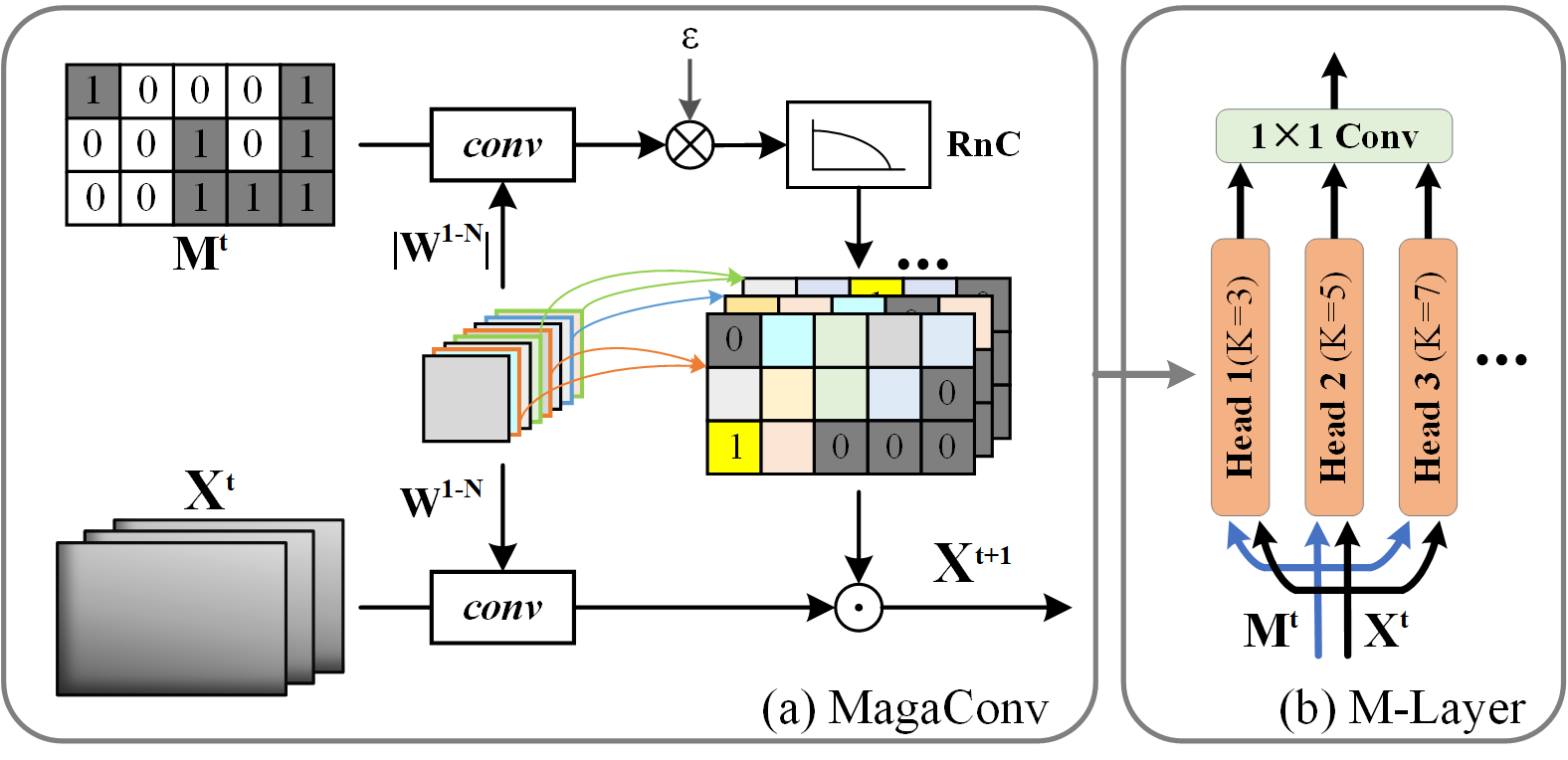}
  \caption{Details of the MagaConv and M-Layer. Each M-Layer consists of multiple MagaConv heads to facilitate feature extraction using diverse kernel sizes. Specifically, we implement three parallel heads with kernel sizes of 3, 5, and 7 for practical application.}
  \label{fig:mag}
\end{figure}

In tackling the challenges posed by invalid pixels during convolution operations, the introduction of PConv partially mitigates their negative effects. To address these issues more effectively, we introduce the MagaConv operation. It employs a convolution kernel selection mechanism to handle invalid patterns without compromising the essential parameters, achieved through the utilization of masks to regulate the convolution process.

\textbf{MagaConv Operation}. Considering a raw depth map $X^t \in \Re^{ h \times w}$ and a vanilla convolution kernel $W$ with the size of $k \times k$ that processes a group of pixels. The output $O^{Conv}$ at the position $(i, j)$ can be defined as follows:

\begin{equation}
    O_{(i, j)}^{Conv} = \sum_{m = -k}^{k} \sum_{n = -k}^{k} W_{(i+m, j+n)} * X_{(i+m, j+n)} .
    \label{eq:VConv}
\end{equation}

Then, denote $M \in \Re^{ h \times w}$ as the corresponding mask of the $X$, in which 1 of the mask map represents a missing depth pixel. It can be defined as follows:

\begin{equation}
    M_{i, j} = \left\{
    \begin{array}{ccc}
    1 & \text{if} & X_{i, j} \leq 0 \\
    0 & \text{if} & X_{i, j} > 0
    \end{array}
    \right.
    .
    \label{eq:mask}
\end{equation}

This mask can be used to mark invalid pixels and then measure the suitability of each pixel within the receptive fields of $W$. Specifically, we suppose that the larger absolute parameter in $W$ is likely to be an important reference. If the missing value is related to that parameter, the output of the convolutional kernel becomes less reliable at that position. Based on the observation, we adopt the same convolution kernel with absolute parameters $|W|$ to conduct convolution operation with the mask $M$, quantitatively measuring the suitability in a specific position. This operation is defined as follows:

\begin{equation}
    O_{i, j}^{Mask} = \epsilon * \sum_{m = -k}^{k} \sum_{n = -k}^{k} | W_{(i+m, j+n)} | * M_{(i+m, j+n)} , 
    \label{eq:VConv_mask}
\end{equation}

\noindent where $O_{i, j}^{Mask}$ indicates the convolution kernel is unsuitable at the position $(i, j)$ when it encounters invalid areas. $\epsilon \in (0, 1)$ is a learnable parameter used to normalize $O_{i, j}^{mask}$, enhancing the training robustness. Notably, the mask $M$ is replicated along the channel axis before the convolution operation, ensuring that its shape remains the same with the depth features. The value $O_{i, j}^{Mask}$ can be transformed into a penalty term, effectively punishing convolutional kernels that heavily rely on invalid areas. The penalty term is defined as follows:

\begin{equation}
    X_{i, j}^{t+1} = RnC(O_{i, j}^{Mask}) \otimes O_{i, j}^{Conv} , 
    \label{eq:mask_conv}
\end{equation}

\noindent where $\otimes$ denotes element-wise multiplication, and the RnC (Reverse-and-Cut) is an activation function we proposed, which is defined as follows:

\begin{equation}
    RnC(x) = \left[ ReLU (e^{-x} - 0.5) \right] \times 2, x > 0 .
    \label{eq:act_mask}
\end{equation}

The RnC function acts as a control mechanism, modulating the convolution process based on the suitability of the key parameters at each position. It is evident that $RnC(x) \in [0, 1]$. Higher values indicate MagaConv behaves more like vanilla convolution, while lower values lead it to act more like Partial Convolution. Additionally, similar to other activation functions, RnC also incorporates a non-linear and threshold-based activation during guidance, enabling convolution kernels to learn complex relationships.

\textbf{Mask Update}. To provide particular instructions for each MagaConv-Layer at different scales, masks are updated in every block and layer according to distinct rules. In layer $l$ of block $b$, the mask updating rule is defined as follows:

\begin{equation}
    \begin{split}
        (1-M(b, l+1)) = MP((1-M(b, l)), s=1) \\
        (1-M(b+1, l)) = MP((1-M(b, l+2)), s=2)
    \end{split}
    ,
    \label{eq:renew_mask}
\end{equation}

\noindent where $MP$ denotes the MaxPooling2D operation. It is to provide accurate localization of the boundary regions and prevent the convolution process from rapidly filling large holes. $s$ denotes the stride parameter, which is aligned with the stride of the M-Layer. In this manner, a specific mask pixel is updated if surrounding areas contain at least one valid mark, and gradually the mask will become fully valid. 

\subsection{Bi-directional Aligning Projection (Bid-AP)}
\label{subsec:bpfusion}

As previously mentioned, aligning cross-modalities is crucial for effectively integrating color and depth features while filtering out irrelevant aspects like surface appearance or textures. Therefore, we introduce Bid-AP, which aims to identify color features relevant to depth, thereby making precise adjustments to the encoding feature and enriching the overall depth representation.

\begin{figure}[t]
  \centering
  \includegraphics[width=1\linewidth]{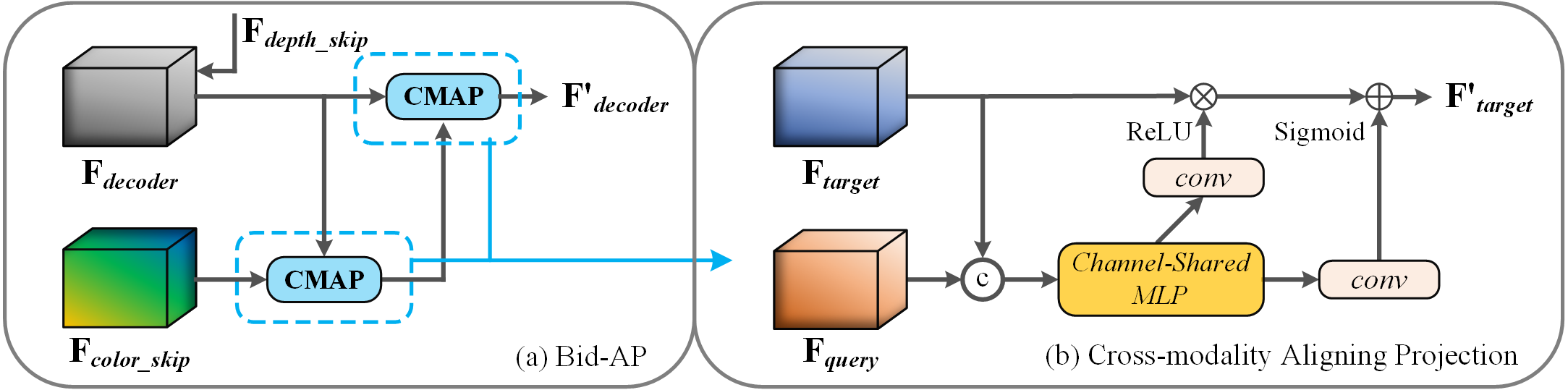}
  \caption{Details of the Bid-AP and CMAP.}
  \label{fig:bid}
\end{figure}

The architecture of the Bid-AP module is illustrated in Fig. \ref{fig:bid} (a). This module aligns the encoder features of the two modalities, $F_{decoder} \& F_{depth\_skip}$ and $F_{color\_skip}$, from the depth ($D$) and color ($C$) respectively. It consists of two parallel streams ($D \to C$ and $C \to D$) to perform a bi-directional information exchange through the CMAP. Initially, after combining $F_{decoder} \& F_{depth\_skip}$ by VConv, $D \to C$ filters out depth-irrelevant features from the color information to emphasize vital aspects like geometric properties. Subsequently, $C \to D$ refines and enriches the representation of depth features. Finally, the decoding features undergo up-sampling via de-convolution operations.

The CMAP is depicted in Fig. \ref{fig:bid} (b). It takes the target feature $F_{target} \in \Re^{h \times w \times c}$ and the query feature $F_{query} \in \Re^{h \times w \times c}$ as inputs, with the target and query representing depth and color interchangeably during the bidirectional fusion. In each step, the CMAP projects $F_{target}$ to align the feature into an enriched pattern. This projecting signal is derived from the concatenated feature $F_{c} = (F_{target}, F_{query})$, generated through an MLP-based spatial-adaptive normalization. Specifically, firstly, $F_{c}$ is fused through a $1 \times 1$ convolution across the channels, resulting in $F'_{c} \in \Re^{h \times w \times c}$. Secondly, it enters a channel-shared MLP layer with two hidden layers to attain a unified representation from a global perspective. The adoption of a channel-shared pattern aims to limit the number of learnable parameters and mitigate the risk of over-fitting. Thirdly, the output embeddings pass through two distinct convolution kernels that produce two modulated signals: $\gamma \in \Re^{h \times w \times c}$ and $\beta \in \Re^{h \times w \times c}$. The overall projection process can be defined as follows:

\begin{equation}
    F'_{target} = ReLU(\gamma) \otimes F_{target} + Sigmoid(\beta), 
    \label{eq:cms}
\end{equation}

\noindent where $\otimes$ denotes element-wise multiplication. 

\begin{figure}[t]
  \centering
  \includegraphics[width=1\linewidth]{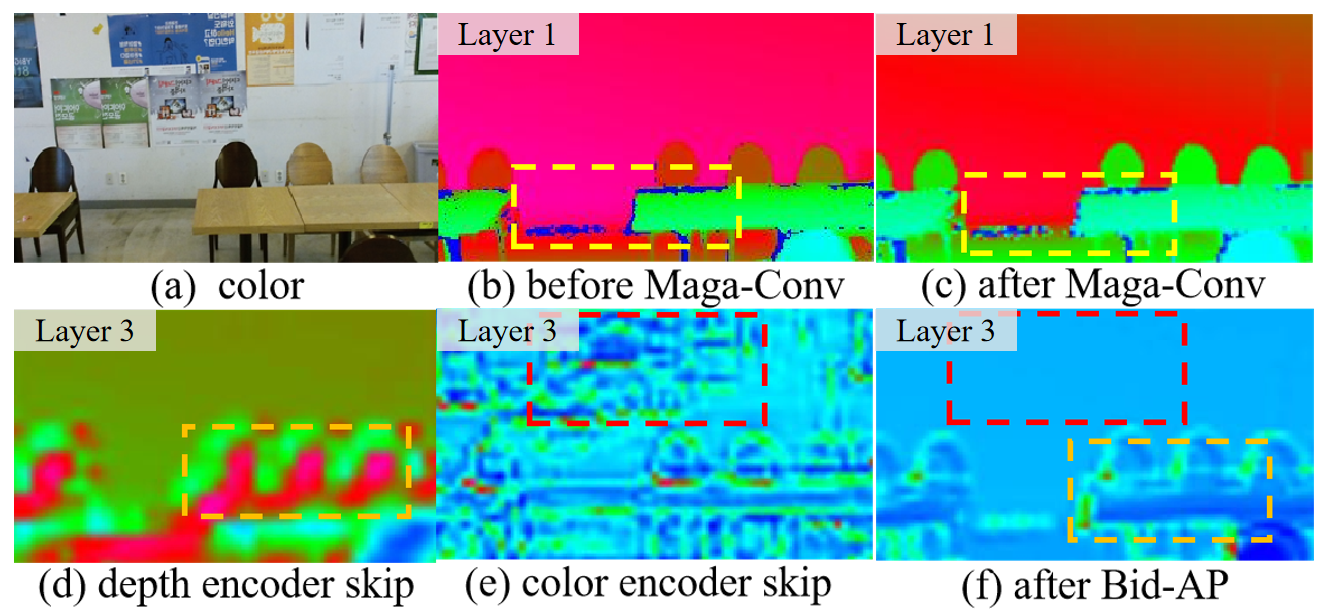}
  \caption{Visualizations of typical features to demonstrate effectiveness on DIML dataset. The missing area in (b) shrinks to (c) after a MagaConv. The green "coarsely complete depth" with unclear boundaries in (d) and the red"depth-irrelevant" features in (e) disappeared after being combined by a Bid-AP (f).}
  \label{fig:visual}
\end{figure}

In general, the Bid-AP module achieves thorough fusion as Fig. \ref{fig:visual} through three key advantages. \textbf{(i) Adaptive Feature selection:} The Bid-AP is to align the features from two modalities, including coarsely complete depth and color features. This alignment surpasses direct feature fusion, representing a learnable selection process that emphasizes the most informative elements from each modality. By avoiding the negative impact caused by directly concatenating both features or introducing depth-irrelevant features, Bid-AP realizes a learnable selection process that emphasizes the most informative elements from each modality. \textbf{(ii) Bi-directional Aligning:} The $D \to C$ acts as a filtering mechanism, converting color skip features into essential features like outlines and semantics. Conversely, $C \to D$ enriches the target features with necessary attributes without overwhelming them with irrelevant contexts from other modalities. \textbf{(iii) Global Perspective with Limited Resources} The CMAP module within the Bid-AP facilitates a global perspective on feature alignment while being resource-efficient. The spatial-channel attention mechanism, implemented via the channel-shared MLP, adapts to crucial contexts across every position, allowing for effective interaction through customized normalization parameters. Compared to other fusion mechanisms like cross-attention, our model achieves efficient alignment with less reliance on training data. 

\subsection{Loss}

It is worth noting that depth maps often contain crucial boundary information that may not be effectively captured by the Mean Squared Error (MSE). Therefore, we employed Structure-Consistence (SC) loss function to address this limitation. The Structure-Consistence loss function can be formulated as follows:

\begin{equation}
    \mathcal{L}_{sc} = \frac{1}{N} \sum_{i=1}^N \left| \nabla D_{pred}^{(i)} - \nabla D_{gt}^{(i)} \right |_2^2 . 
    \label{eq:loss}
\end{equation}

$\mathcal{L}_{sc}$ represents the structure-consistence loss, $N$ is the number of samples in the training process, $D{pred}^{(i)}$ is the predicted depth map for the $i$-th sample, $D_{gt}^{(i)}$ is the corresponding ground truth depth map, $\nabla$ denotes the Laplacian operator for extracting edge information, and $\left| \cdot \right|_2^2$ denotes the squared Euclidean norm.

The overall loss function is given by:

\begin{equation}
\mathcal{L}_{all} = \mathcal{L}_{mse} + \mathcal{L}_{sc} . 
\label{eq:loss_all}
\end{equation}

By incorporating the SC loss with MSE loss, the model is encouraged to minimize not only the pixel-wise depth errors but also to preserve the structural integrity and edge information. This leads to more visually accurate and detailed depth completion results.

\section{Experiments}

\begin{table*}[ht]
    \centering
    \setlength{\tabcolsep}{2pt}
    \caption{Ablation study results for different schemes of the pipeline on the NYU-Depth V2 datasets. RMSE is the main metric.}
    \scalebox{0.8}{
    \begin{tabular}{c|cccc|cccc|cc|ccc}
        \hline
        Scheme & MagaConv & M-Layer & PConv & GConv & Bi-direction & CMAP & MLP-based & Concat & $\mathcal{L}_{mse}$ & $\mathcal{L}_{sc}$ & RMSE$\downarrow$ & Rel$\downarrow$ & $\delta_{1.10}\uparrow$ \\
        \hline\hline
        A (baseline) & - & - & - & - & - & - & - & - & \ding{51} & \ding{51} & 0.188 & 0.028 & 95.6\\
        \hline
        B (w/ MagaConv) & \ding{51} & \ding{51} & - & - & - & - & - & - & \ding{51} & \ding{51} & \textbf{0.109} & \textbf{0.015} & \textbf{97.3}\\
        C & \ding{51} & - & - & - & - & - & - & - & \ding{51} & \ding{51} & 0.114 & 0.016 & 97.0\\
        D & - & \ding{51} & \ding{51} & - & - & - & - & - & \ding{51} & \ding{51} & 0.134 & 0.018 & 96.4\\
        E & - & \ding{51} & - & \ding{51} & - & - & - & - & \ding{51} & \ding{51} & 0.127 & 0.017 & 96.7\\
        \hline
        F (w/ Bid-AP) & - & - & - & - & \ding{51} & \ding{51} & \ding{51} & - & \ding{51} & \ding{51} & \textbf{0.113} & \textbf{0.016} & \textbf{97.1}\\
        G & - & - & - & - & - & \ding{51} & \ding{51} & - & \ding{51} & \ding{51} & 0.139 & 0.018 & 96.1\\
        H & - & - & - & - & \ding{51} & \ding{51} & - & - & \ding{51} & \ding{51} & 0.125 & 0.017 & 96.8\\
        I & - & - & - & - & \ding{51} & - & - & \ding{51} & \ding{51} & \ding{51} & 0.135 & 0.018 & 96.4\\
        \hline
        J & \ding{51} & \ding{51} & - & - & \ding{51} & \ding{51} & \ding{51} & - & \ding{51} & - & 0.087 & 0.012 & 98.2\\
        K (complete) & \ding{51} & \ding{51} & - & - & \ding{51} & \ding{51} & \ding{51} & - & \ding{51} & \ding{51} & \textbf{0.083} & \textbf{0.011} & \textbf{98.7}\\
        \hline
    \end{tabular}}
\label{tab:ablation}
\end{table*}

\subsection{Experimental Setup}

We conducted comprehensive experiments on three popular benchmark datasets: NYU-Depth V2, DIML, and SUN RGB-D to validate the performance of the model.

\textbf{NYU-Depth V2} \cite{10.1007/978-3-642-33715-4_54} is the most authoritative and widely used benchmark dataset for depth image completion, which contains 408,473 images collected in 464 different indoor scenes, and 1449 officially labeled images for evaluation.

\textbf{DIML} \cite{DBLP:journals/corr/abs-1904-10230} This dataset includes images with typical edge shadows and irregular holes, providing a robust evaluation benchmark for assessing the adaptability of our model to various invalid patterns. We utilize 2000 pairs of labeled samples from the indoor part of the datasets according to the official split. 

\textbf{SUN RGB-D} \cite{Song_2015_CVPR} is an extensive dataset comprising 10,335 densely captured RGB-D images obtained from four different sensors. The dataset covers 19 primary scene categories, providing a diverse range of scenes for evaluation. Following the default protocol, we partitioned the datasets into 4,845 images for training and 4,659 ones for testing.

\textbf{Metrics.} The evaluation of indoor depth completion results is based on three criteria: Root Mean Squared Error (RMSE), Relative Error (Rel), and Threshold Accuracy ($\delta_t$) with thresholds $t=1.10, 1.25, 1.25^2, 1.25^3$. 

\begin{figure}[t]
\centering
  \centering
  \includegraphics[width=0.8\linewidth]{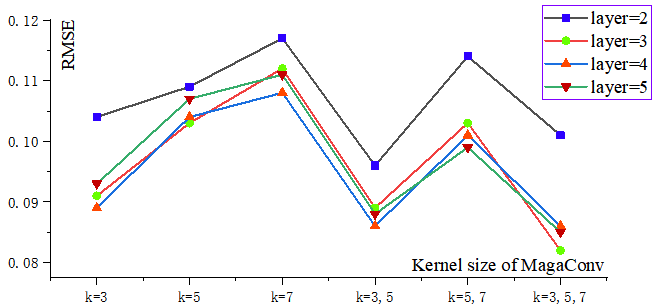}
  \caption{Analyzing the performance across various configurations involving different numbers of down-sampling layers and sets of kernel sizes for MagaConv. The best performance is observed with kernel sizes of 3, 5, and 7 alongside 3 down-sampling layers.}
\label{fig:ks}
\end{figure}

\textbf{Implementation Details.} Our model was implemented using the PyTorch framework and trained on NVIDIA GTX 2080ti GPU for a total of 100 epochs. We adopted the SGD optimizer for training, with a momentum term of 0.95 and a weight decay term of $10^{-4}$. The initial learning rate was set to $1 \times 10^{-3}$ and was halved during the plateau period. The model was trained using end-to-end training methodology.

\subsection{Ablation Studies}

To optimize the proposed framework and evaluate its performance, ablation experiments were conducted on the NYU-Depth V2 datasets. At first, a baseline model (Scheme A) was constructed to resemble the proposed framework, retaining the encoder-decoder architecture but using vanilla convolution in place of the MagaConv operation, and employing direct concatenation of depth and color features at the bottleneck instead of the Bid-AP module. Based on this baseline, three categories with nine protocols (Schemes B to K) were designed by combining different configurations for each module. Schemes B–E evaluated the effectiveness of MagaConv, Schemes F–I examined the impact of the Bid-AP module, and Schemes J and K explored the influence of different loss functions. The details of these protocols are summarized in Tab. \ref{tab:ablation}.

\begin{figure*}[t]
\centering
  \centering
  \includegraphics[width=0.95\linewidth]{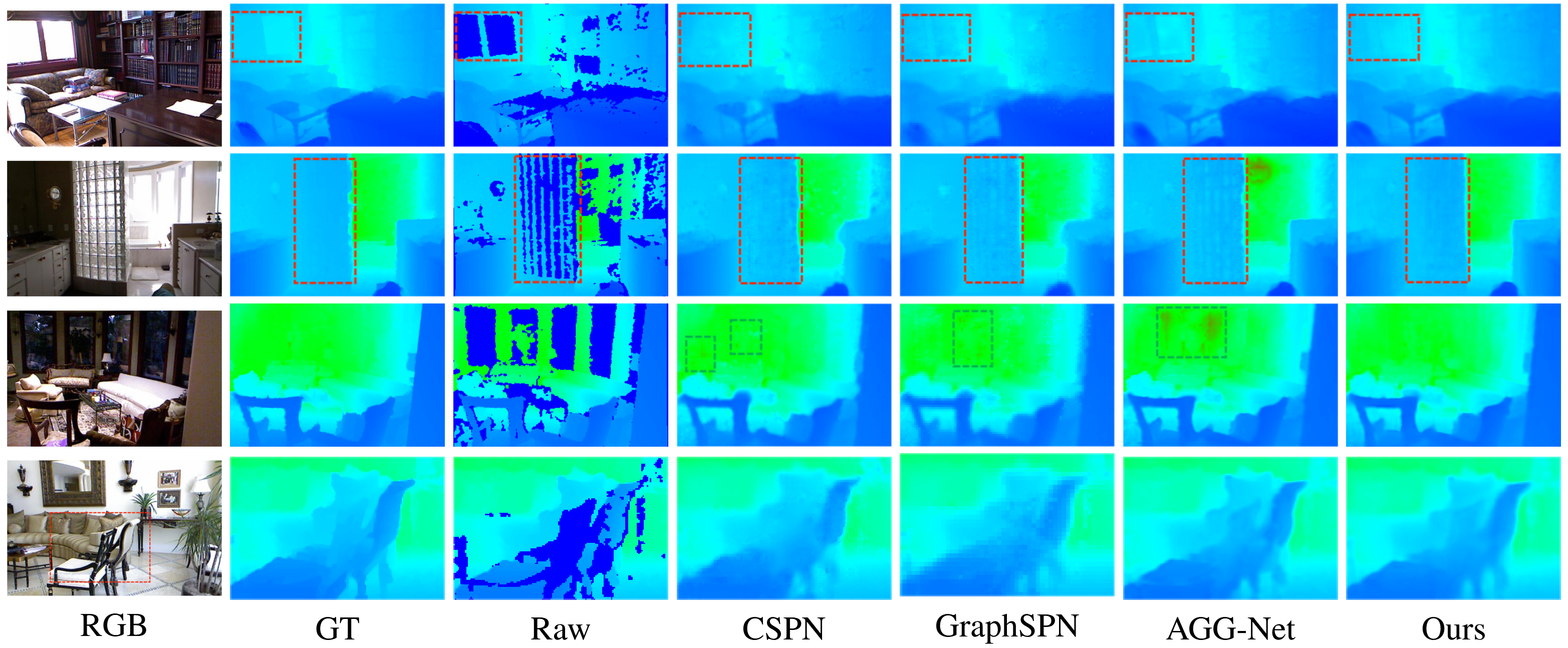}
  \caption{Depth completion comparison results with different methods on NYU-Depth V2.}
\label{fig:comparison}
\end{figure*}

\textbf{(i) On MagaConv.} In the first group, we investigated the impact of different depth encoding methods while maintaining the remaining settings identical to the baseline. Scheme B incorporated the complete MagaConv module, which demonstrated improved performance compared to the baseline (0.109 v.s. 0.188 RMSE). Scheme C involved a modification where the three parallel MagaConv in each M-layer were replaced with a single MagaConv with a $5 \times 5$ kernel. This results in a minor performance decrease across all metrics (0.114 v.s. 0.109 RMSE), and indicates that the M-Layer is essential in capturing multi-scale features. In Scheme D and E, substituting the MagaConv with Partial convolution and Gated convolution led to a performance decline across all metrics (0.134 and 0.127 v.s. 0.109 RMSE), indicating that MagaConv indeed enhances the depth features' reliability. Furthermore, the parameters' ablation experiments are depicted in Fig. \ref{fig:ks}, and the visualization before and after MagaConv is provided in Fig. \ref{fig:visual} (b) and (c). This also suggests that the MagaConv module effectively filters out invalid features, resulting in a more reliable feature representation.

\textbf{(ii) On Bid-AP.} In the second group, various fusion schemes were integrated into the baseline. Scheme F, featuring our novel Bid-AP module, displayed a significant performance boost compared to Scheme A (0.113 v.s. 0.188 RMSE), showcasing that the alignment of depth and color features by Bid-AP enhances depth map reconstruction accuracy. In Scheme G, replacing the bi-directional module with a unidirectional approach ($C \to D$) resulted in a performance decline (0.139 v.s. 0.113 RMSE), implying that Bid-Aligning aids in the filtering and fusion process. In scheme H, the channel-shared MLP within the CMAP was removed to facilitate a localized fusion process. This resulted in a noticeable decrease across all metrics (0.125 v.s. 0.113 RMSE), indicating that the global perspective plays an important role in the comprehensive alignment features. In scheme I, conventional concatenation and convolution structures replaced CMAP. The findings suggest that the fusion strategy centered around normalization, a core concept employed by CMAP, proves to be more effective (0.135 v.s. 0.113 RMSE). Additionally, the visualization of features related to Bid-AP is presented in Fig. \ref{fig:visual} (d-f), demonstrating that depth-irrelevant features are effectively filtered out in the color encoder and seamlessly fused with the extracted depth features, further validating its effectiveness.

\textbf{(iii) On loss function.} In the last group, two different settings of loss functions are evaluated. Scheme K, which integrated both the losses significantly outperforms scheme J which only employed the MSE loss (0.083 v.s. 0.087 RMSE). It demonstrates the effectiveness of integrating the structure-consistency loss into our approach, leading to enhanced performance in depth map completion.

\subsection{Comparison to State-of-the-art}

\begin{table}[t]
  \centering
  \setlength{\tabcolsep}{2pt}
  \caption{Quantitative evaluation on NYU-Depth V2 dataset.}
  \scalebox{0.85}{
  \begin{tabular}{c|c|c|c|ccc}
    \hline 
     Method & Params & RMSE$\downarrow$ & Rel$\downarrow$ & $\delta_{1.25}\uparrow$ & $\delta_{1.25^2}\uparrow$ & $\delta_{1.25^3}\uparrow$ \\
    \hline\hline
    CSPN++ & 17.4 M & 0.173 & 0.02 & 96.3 & 98.6 & 99.5 \\
    NLSPN & 25.8 M & 0.153 & 0.015 & 98.6 & 99.6 & 99.9\\
    RDF-GAN & - M & 0.139 & 0.013 & 98.7 & 99.6 & 99.9\\
    GraphCSPN &  - M & 0.133 & 0.012 & 98.8 & 99.7 & 99.9\\
    AGG-Net & 129.1 M & 0.092 & 0.014 & 99.4 & 99.9 & 100.0\\
    CFormer & 146.7 M & 0.091 & 0.012 & 99.6 & 99.9 & 100.0\\
    TPVD  & 31.2 M & 0.086 & 0.010 & 99.7 & 99.9 & 100.0\\
    \textbf{Ours} & 30.1 M & \textbf{0.083} & 0.011 & \textbf{99.7} & \textbf{99.9} & \textbf{100.0}\\
    \hline
  \end{tabular}
}
\label{tab:nyu}
\end{table}

\begin{table}[t]
  \centering
  \setlength{\tabcolsep}{2pt}
  \caption{Quantitative comparison results with competing methods on DIML and SUN RGB-D datasets.}
  \scalebox{0.85}{
  \begin{tabular}{c|c|c|c|ccc}
    \hline 
     Benchmark & Method & RMSE$\downarrow$ & Rel$\downarrow$ & $\delta_{1.25}\uparrow$ & $\delta_{1.25^2}\uparrow$ & $\delta_{1.25^3}\uparrow$ \\
    \hline\hline
    \multirow{6}{*}{\makecell[c]{DIML}} 
    ~ & CSPN++  & 0.162 & 0.033 & 96.1 & 98.7 & 99.6\\
    ~ & DfuseNet  & 0.143 & 0.023 & 98.4 & 99.4 & 99.9\\
    ~ & DM-LRN  & 0.149 & 0.015 & 99.0 & 99.6 & 99.9\\
    ~ & NLSPN  & 0.114 & 0.013 & 99.2 & 99.7 & 99.9\\
    ~ & AGG-Net  & 0.086 & 0.011 & 99.6 & 99.9 & 100.0\\
    ~ & \textbf{Ours} & \textbf{0.060} & \textbf{0.010} & \textbf{99.8} & \textbf{99.9} & \textbf{100.0}\\
    \hline
    \multirow{5}{*}{\makecell[c]{SUN RGBD}} 
    ~ & CSPN++  & 0.295 & 0.137 & 95.6 & 97.5 & 98.4\\
    ~ & NLSPN  & 0.267 & 0.063 & 97.3 & 98.1 & 98.5\\
    ~ & RDF-GAN  & 0.255 & 0.059 & 96.9 & 98.4 & 99.0\\
    ~ & AGG-Net  & 0.202 & 0.038 & 98.5 & 99.0 & 99.4\\
    ~ & \textbf{Ours} & \textbf{0.197} & 0.039 & \textbf{98.5} & \textbf{99.2} & \textbf{99.6}\\
    \hline
  \end{tabular}
  }

  \label{tab:comparison}
\end{table}

To evaluate the performance of our proposed model, we conducted comparative experiments against state-of-the-art depth completion methods. 

\textbf{On NYU-Depth V2.} The quantitative comparison results with other state-of-the-art methods \cite{cheng2020cspn++, NLSPN, RDF_GAN, graphcspn, aggnet, completionformer, yan2024tri} on NYU-Depth V2 datasets are shown in Tab. \ref{tab:comparison}. Our model performs well across all metrics while maintaining relatively low parameter counts. Visual results in Fig. \ref{fig:comparison} further emphasize its effectiveness, demonstrating clearer details in challenging scenarios. For instance, our method recovers missing window regions with greater clarity in the first two rows than competitors. In the bottom row, it captures finer details, such as sharper chair edges, and avoids the unrealistic artifacts observed in other methods.
Additionally, efficiency tests on a single RTX 3090 GPU at 192×320 resolution show our model achieves 101.7 GFlops, 41ms runtime, and 24.4 FPS. These results demonstrate its suitability for real-time applications with reduced computational demand and improved efficiency.

\textbf{On DIML and SUN RGB-D.} Regarding the dataset DIML, our model is also compared to state-of-the-art methods \cite{cheng2020cspn++, dfusenet, DM_LRN, NLSPN, aggnet}. Our model outperforms these competing methods in all three metrics, with a remarkable 20\% improvement in RMSE. For the datasets SUN RGB-D, our model is evaluated against comparative methods including \cite{cheng2020cspn++, NLSPN, RDF_GAN, aggnet}, and also achieved competing performance.

\section{Conclusion}

In our research, we introduced a novel method for indoor depth completion by integrating the MagaConv and Bid-AP modules to improve accuracy and reliability. The MagaConv architecture strategically selects convolution kernels based on updated masks, facilitating precise depth feature extraction. Bid-AP aligns features from two modalities using a global bi-directional projection approach. Our model outperformed current state-of-the-art methods on datasets with relatively low parameter counts. Looking ahead, while our focus lies on indoor depth completion with TOF cameras rather than sparse depth completion in the future, we aim to extend this innovative technique to diverse applications, to better contribute to further tasks.

\section{Acknowledgments}
This research was supported by the National Natural Science Foundation of China (62202087, 62173083, U22A2063), Guangdong Basic and Applied Basic Research Foundation 2024A1515010244, Fundamental Research Funds for the Central Universities (N2404008, N2404011), and the 111 Project B16009.

\bibliography{aaai25}

\end{document}